\documentclass{amsart}
\usepackage{natbib}
\usepackage{graphicx}
\usepackage{mathptmx}
\usepackage{latexsym}
\usepackage{url}

\newcommand{\bof}{Bag of Features} 
\newcommand{\ic}{image classification}
\newcommand{\ir}{image retrieval}

\begin{document}

\title{Introduction to the Bag of Features Paradigm for Image Classification and Retrieval}
\author{Stephen O'Hara}
\author{Bruce A. Draper}
\address{
Colorado State University \\
Computer Science Department \\
Fort Collins, CO 80523
}
\email{ \{svohara, draper\}@cs.colostate.edu}
\date{July 2010}

\begin{abstract}
The past decade has seen the growing popularity of \bof{} (BoF) approaches to many computer vision tasks, including image classification, video search, robot localization, and texture recognition. Part of the appeal is simplicity. BoF methods are based on orderless collections of quantized local image descriptors; they discard spatial information and are therefore conceptually and computationally simpler than many alternative methods. Despite this, or perhaps because of this, BoF-based systems have set new performance standards on popular image classification benchmarks and have achieved scalability breakthroughs in image retrieval. This paper presents an introduction to BoF image representations, describes critical design choices, and surveys the BoF literature. Emphasis is placed on recent techniques that mitigate quantization errors, improve feature detection, and speed up image retrieval. At the same time, unresolved issues and fundamental challenges are raised. Among the unresolved issues are determining the best techniques for sampling images, describing local image features, and evaluating system performance. Among the more fundamental challenges are how and whether BoF methods can contribute to localizing objects in complex images, or to associating high-level semantics with natural images.  This survey should be useful both for introducing new investigators to the field and for providing existing researchers with a consolidated reference to related work.
\end{abstract}
\maketitle
%
\section{Introduction}
\label{section:Intro}
The past decade has seen the rise of the \bof{} approach in computer vision. \bof{} (BoF) methods have been applied to \ic{}, object detection, \ir{}, and even visual localization for robots. BoF approaches are characterized by the use of an orderless collection of image features. Lacking any structure or spatial information, it is perhaps surprising that this choice of image representation would be powerful enough to match or exceed state-of-the-art performance in many of the applications to which it has been applied. Due to its simplicity and performance, the \bof{} approach has become well-established in the field. This paper seeks to characterize the \bof{} paradigm, providing a survey of relevant literature and a discussion of open research issues and challenges. We focus on the application of BoF to weakly supervised \ic{} and unsupervised \ir{} tasks.

This survey is of interest to the computer vision community for three reasons. First, \bof{} methods work. As discussed below, BoF-based systems have demonstrated comparable or better results than other approaches for \ic{} and \ir{}, while being computationally cheaper and conceptually simpler. Second, the BoF approach has appeared under different names in several seemingly unrelated branches of the literature. Besides the computer vision literature, where the term ``\bof{}" was coined, closely related approaches appear in the literature on biological modeling, texture analysis, and robot localization. As a result, there is more work on BoF than many researchers may be aware of. Finally, the \bof{} approach is a multi-step process, with each step presenting many options. Many plausible combinations have never been tried. We hope to contribute to future advances in the field through this survey by mapping out the space of BoF algorithms, recording what is known about the steps and how they interact, and identifying remaining research opportunities.

Although there is no contemporary survey on \bof{} methods, related surveys include Frintrop's survey of computational visual saliency \citep{frintrop_computational_2010} and Datta's overview of Content-Based Image Retrieval (CBIR) techniques \citep{datta_image_2008}. Frintrop reviews techniques that share similarities to the feature extraction stage of BoF methods, as discussed in Section \ref{section:FDE} of this report. Datta's CBIR survey discusses BoF-based \ir{} to some extent, but we present a broader survey of BoF techniques with more details on state-of-the-art methods and specific mechanisms that have been employed to improve query results and speed.

This paper is organized as follows. Section \ref{section:BOF} provides an overview of the \bof{} image representation. Section \ref{section:FDE} provides details on the feature detection and extraction techniques commonly employed in BoF representations. Vector Quantization is an important aspect of the BoF approach, and so Section \ref{section:quantization} discusses the quantization challenges and how the BoF community has attempted to address them. Two popular applications of BoF representations include \ic{} and \ir{}, which are presented in Sections \ref{section:IC} and \ref{section:IR}, respectively. Section \ref{section:eval} looks at the evaluation of BoF methods, including common performance measures, data sets, and the challenges involved with comparative evaluation. Although BoF methods have shown promising performance in a number of tasks, there remain open issues and inherent limitations. We explore a few of these in Section \ref{section:challenges}. Section \ref{section:conclusion} has our concluding remarks.

\section{Bag of Features Image Representation}
\label{section:BOF}
A \bof{} method is one that represents images as orderless collections of local features. The name comes from the Bag of Words representation used in textual information retrieval. This section provides an explanation of the \bof{} image representation, focusing on the high-level process independent of the application. More sophisticated variations and other implementation details are discussed later in this report.

There are two common perspectives for explaining the BoF image representation. The first is by analogy to the Bag of Words representation. With Bag of Words, one represents a document as a normalized histogram of word counts. Commonly, one counts all the words from a dictionary that appear in the document. This dictionary may exclude certain non-informative words such as articles (like ``the''), and it may have a single term to represent a set of synonyms. The \textit{term vector} that represents the document is a sparse vector where each element is a term in the dictionary and the value of that element is the number of times the term appears in the document divided by the total number of dictionary words in the document (and thus, it is also a normalized histogram over the terms). The term vector is the Bag of Words document representation -- called a ``bag'' because all ordering of the words in the document have been lost.

The \bof{} image representation is analogous. A visual vocabulary is constructed to represent the dictionary by clustering features extracted from a set of training images. The image features represent local areas of the image, just as words are local features of a document. Clustering is required so that a discrete vocabulary can be generated from millions (or billions) of local features sampled from the training data. Each feature cluster is a visual word. Given a novel image, features are detected and assigned to their nearest matching terms (cluster centers) from the visual vocabulary. The term vector is then simply the normalized histogram of the quantized features detected in the image.

The second way to explain the BoF image representation is from a codebook perspective. Features are extracted from training images and vector quantized to develop a visual codebook. A novel image's features are assigned the nearest code in the codebook. The image is reduced to the set of codes it contains, represented as a histogram. The normalized histogram of codes is exactly the same as the normalized histogram of visual words, yet is motivated from a different point of view. Both ``codebook'' and ``visual vocabulary'' terminology is present in the surveyed literature.

The BoF term vector is a compact representation of an image which discards large-scale spatial information and the relative locations, scales, and orientations of the features. A contemporary large-scale BoF-based \ir{} system might have a dictionary of 100,000 visual words and 5,000 features extracted per image. Thus in an image where there are no duplicate visual words (unusual), the term vector will have 95\% of its elements as zeros. The strong sparsity of term vectors allows for efficient indexing schemes and other performance improvements, as discussed in later sections.

\begin{figure*}[ht]
  \centering
  \includegraphics[width=0.85\textwidth]{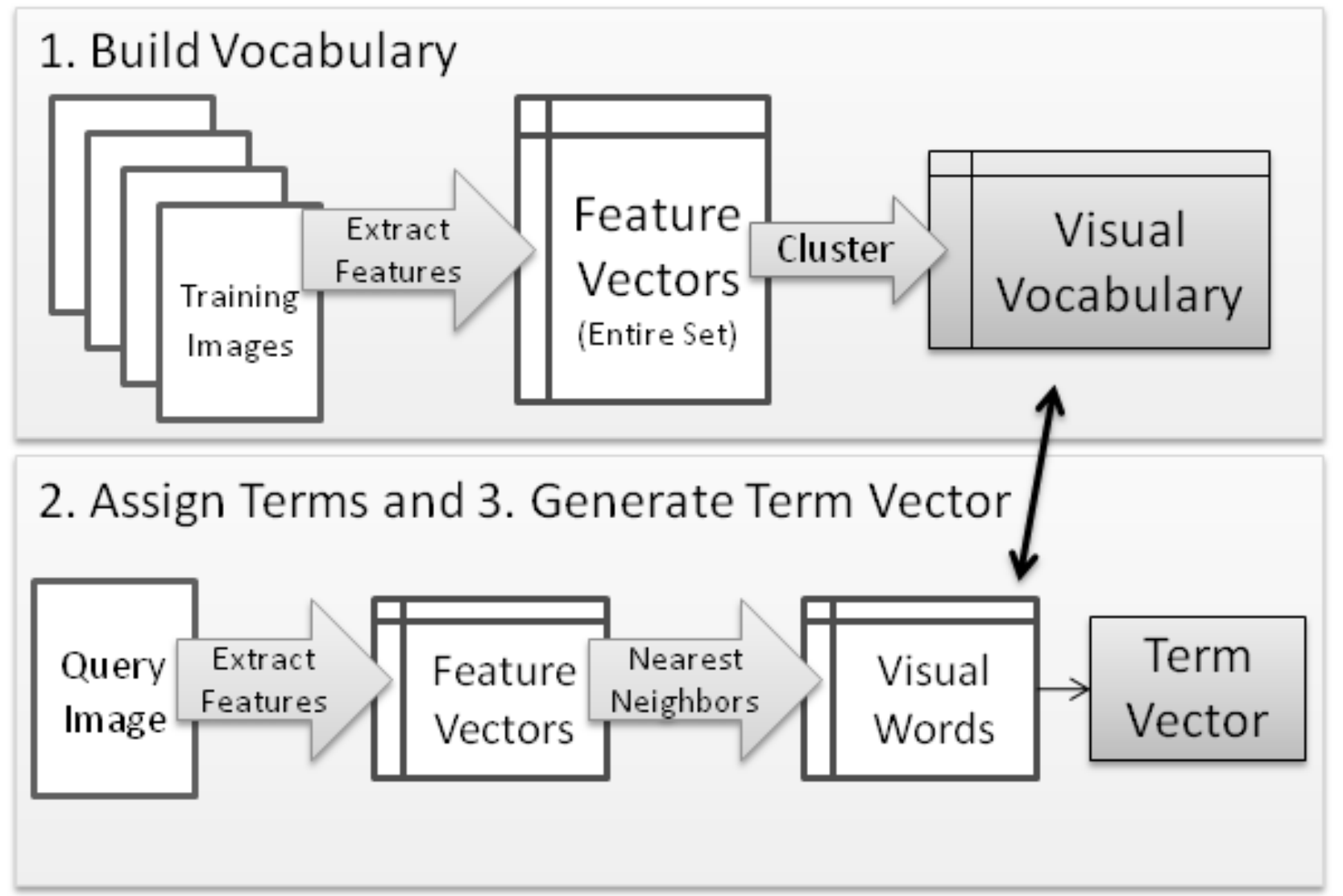}
  \caption{Process for \bof{} Image Representation}
  \label{fig:BoF Process}
\end{figure*}

At a high level, the procedure for generating a \bof{} image representation is shown in Figure \ref{fig:BoF Process} and summarized as follows:
\begin{enumerate}
	\item \textbf{Build Vocabulary}: Extract features from all images in a training set. Vector quantize, or cluster, these features into a ``visual vocabulary,'' where each cluster represents a ``visual word'' or ``term.'' In some works, the vocabulary is called the ``visual codebook.'' Terms in the vocabulary are the codes in the codebook.
	\item \textbf{Assign Terms}: Extract features from a novel image. Use Nearest Neighbors or a related strategy to assign the features to the closest terms in the vocabulary.
	\item \textbf{Generate Term Vector}: Record the counts of each term that appears in the image to create a normalized histogram representing a ``term vector.'' This term vector is the \bof{} representation of the image. Term vectors may also be represented in ways other than simple term frequency, as discussed later.
\end{enumerate}

There are a number of design choices involved at each step in the BoF representation. One key decision involves the choice of feature detection and representation. Many use an interest point operator, such as the Harris-Affine detector \citep{mikolajczyk_comparison_2005} or the Maximally Stable Extremal Regions (MSER) detector \citep{matas_robust_2004}. At every interest point, often a few thousand per image, a high-dimensional feature vector is used to describe the local image patch. Lowe's 128-dimension SIFT descriptor is a popular choice \citep{lowe_object_1999}. Other choices of feature detection and representation are discussed in Section \ref{section:FDE}.

Another pair of design choices involve the method of vector quantization used to generate the vocabulary and the distance measure used to assign features to cluster centers. A distance measure is also required when comparing two term vectors for similarity (as is done with \ir{}), but this measure operates in the term vector space as opposed to the feature space. Quantization issues and the choice of distance measure can impact term assignment and similarity scoring. These issues are discussed in Section \ref{section:quantization}.

\section{Feature Detection and Representation}
\label{section:FDE}
If you are going to represent an image as a \bof{}, the feature had better be good! This may be easier said than done. There are many possible approaches to sampling image features, including ``Interest Point Operators," ``Visual Saliency," and random or deterministic grid sampling. Further, the definition of what makes a good feature may be application-dependant, and a universally-accepted measure of fitness for localized features has yet to be developed.

\subsection{Feature Detection}
\label{sect:f_detection}
Feature detection is the process of deciding where and at what scale to sample an image. The output of feature detection is a set of keypoints that specify locations in the image with corresponding scales and orientations. These keypoints are distinct from feature descriptors, which encode information from the pixels in the neighborhood of the keypoints. Thus, feature detection is a separate process from feature representation in BoF approaches. Feature descriptors are presented in Section \ref{sect:f_descriptors}.

There is a substantial body of literature that focuses on detecting the location and extent of good features from at least two different sub-fields of computer vision. The first developed from the goal of finding keypoints useful for image registration that are stable under minor affine and photometric transformations. These feature detection methods are referred to as \textit{Interest Point Operators}. The second group detects features based on computational models of the human visual attention system. These methods are concerned with finding locations in images that are visually salient. In this case, fitness is often measured by how well the computational methods predict human eye fixations recorded by an eye tracker. In the following two subsections, we discuss both approaches to feature detection.

Finally, there is research that suggests generating keypoints by sampling the images using a grid or pyramid structure, or even by random sampling. Deterministic and random sampling approaches are discussed in this section as well, below.

\subsubsection{Interest Point Operators}

While there are many variations, an interest point operator typically detects keypoints using scale space representations of images. A scale space represents the image at multiple resolutions, and is generated by convolving the image with a set of guassian kernels spanning a range of $\sigma$ values. The result is a data structure which is, among other things, a convenient way to efficiently apply image processing operations at multiple scales. For details on scale space representations, see \citet{lindeberg_detecting_1993}. Interest point operators detect locally discriminating features, such as corners, blob-like regions, or curves. Responses to a filter designed to detect these features are located in a three dimensional coordinate space, $(x,y,s)$, where $(x,y)$ is the pixel location and $s$ is the scale. Extremal values for the responses over local $(x,y,s)$ neighborhoods are identified as interest points.

Perhaps the most popular keypoint detector is that developed by Lowe \citep{lowe_object_1999}, which employs a Difference-of-Gaussians (DoG) filter for detection. DoG responses can be conveniently computed from a scale space structure, and extremal response values within a local $(x,y,s)$ region used as interest points. Kadir and Brady designed a keypoint detector called Scale Saliency \citep{kadir_saliency_2001} to find regions which have high entropy within a local scale-space region. Another popular keypoint detector, called the Harris-Affine detector \citep{mikolajczyk_scale_2004}, extends the well-known Harris Corner Detector \citep{harris_combined_1988} to a scale space representation with oriented elliptical regions. The Maximally Stable Extremal Regions (MSER) keypoint detector finds elliptical regions based on a watershed process \citep{matas_robust_2004}. These are but a handful of examples of the many interest point operators designed to be robust to small affine and photometric image transformations, with the goal of being able to find the same keypoints in two similar but distinct images.

In practice, the state-of-the-art \bof{} methods tend to use similar feature detectors. Many use either the Harris-Affine or the MSER feature detectors. In large part, this is due to a study in 2005 of affine region detectors \citep{mikolajczyk_comparison_2005}, published as a collaborative effort by many leading researchers in the field. While the conclusions of this study are subject to interpretation, the Harris-Affine and MSER detectors performed well under a variety of situations.

\subsubsection{Visual Saliency}
\bof{} methods often rely on interest point operators to detect the location and scale of localized regions from which image features are extracted. Similarly, many biologically-inspired, or \textit{biomimetic}, computer vision systems use localized regions as well. In the biomimetic computer vision literature, the interest point operator is based on computational models of visual attention. A recent survey by Frintrop et al. \citep{frintrop_computational_2010} provides details on many computational visual attention methods, so we do not reproduce that material here. Our intent is to point out the similarities between the BoF literature, which uses Interest Point Operators, to the visual saliency literature, with the hope that there will be more direct cross-referencing between the two fields in the future.

Itti and Koch proposed a popular model which builds upon a thread of research started by Koch and Ullman in the 1980's \citep{itti_saliency-based_2000,koch_shifts_1985}. The Itti and Koch saliency model, at a high level, looks for extrema in center-surround patterns across several feature channels, such as color opponency, orientation (Gabor filters), and intensity. The center-surround extrema can be detected using Laplacian-of-Gaussian (LoG) or Difference-of-Gaussian (DoG) filters, similar to the models of Lindeberg \citep{lindeberg_detecting_1993} and Lowe \citep{lowe_object_1999}.

Bruce and Tsotsos present an information theoretic approach to saliency \citep{bruce_saliency_2009}, which is conceptually similar to Kadir and Brady's Scale Saliency interest point operator mentioned earlier. A local region is interesting not just based upon a certain pattern or filter response, but also because it differs significantly from the surrounding context of neighboring pixels.

Other computational models of visual saliency continue to be explored in the literature, with an interdisciplinary sub-field that is interested in which computational model best predicts human attentional fixations as measured with an eye-tracker \citep{elazary_interesting_2008,kienzle_center-surround_2009,tatler_visual_2005}. 

\subsubsection{Deterministic and Random Sampling}

A more fundamental question than \textit{which} interest point detector to use, is \textit{whether or not to use an interest point detector at all}. One may characterize the extraction of localized features as an image sampling problem. While keypoint operators are useful for image alignment, it is an open question whether this is the ideal way to sample localized features for image matching or classification.

For example, in the \textit{Video Google} paper \citep{sivic_video_2003}, the authors present a pair of similar images from the movie ``Run Lola, Run'' that fails to match well in a BoF \ir{} task. The images show Lola running down a sidewalk in an urban area. The images have high amounts of low texture concrete -- sidewalk, building facade, and roadway. An illustration of keypoint locations shows that feature detection leaves about half of the images unsampled. While ``uninteresting'' to the feature detectors, the blandness of large portions of the images is in itself potentially highly discriminating.

Mar\'{e}e et al. describe an image classification algorithm featuring random multiscale subwindows and ensembles of randomized decision trees  \citep{maree_random_2005}. While the algorithm is not strictly a BoF approach, it illustrates the efficacy of random sampling. Nowak, Jurie, and Triggs explored sampling strategies for BoF \ic{} in \citep{nowak_sampling_2006}. They show that when using enough samples, random sampling exceeds the performance of interest point operators. They present evidence that the most important factor is the number of patches sampled from the test image, and thus claim dense random sampling is the best strategy.

Spatial Pyramid Matching \citep{lazebnik_beyond_2006} uses SIFT descriptors extracted from a dense grid with a spacing of 8 pixels. K-means clustering is used for constructing the vocabulary. But instead of directly forming the term vector for a given image, the terms are collected in a pyramid of histograms, where the base level is equivalent to the standard BoF representation for the complete image. At each subsequent level in the pyramid, the image is divided into four subregions, in a recursive manner, with each region at each pyramid level having its own histogram (term vector). The distance between two images using this spatial pyramid representation is a weighted histogram intersection function, where weights are largest for the smallest regions. Doing so, Lazebnik captures a degree of location information beyond the standard orderless BoF representation.

\subsection{Feature Descriptors}
\label{sect:f_descriptors}
In addition to determining where and to what extent a feature exists in an image, there is a separate body of research to determine how to represent the neighborhood of pixels near a localized region, called the feature descriptor. The simplest approach is to simply use the pixel intensity values, scaled for the size of the region, or an eigenspace representation thereof. Normalized pixel representations, however, have performed worse than many more sophisticated representations (see \citet{feifei_bayesian_2005,nowak_sampling_2006}, among others) and have largely been abandoned by the BoF research community.

The most popular feature descriptor in the BoF literature is the SIFT (Scale Invariant Feature Transform) descriptor \citep{lowe_distinctive_2004}. In brief, the 128 dimensional SIFT descriptor is a histogram of responses to oriented gradient filters. The responses to 8 gradient orientations at each of 16 cells of a 4x4 grid generate the 128 components of the vector. The histograms in each cell are block-wise normalized. At scale 1, the cells are often 3x3 pixels.

An alternative to the SIFT descriptor that has gained increasing popularity is SURF (Speeded Up Robust Features) \citep{bay_surf:_2006}. The SURF algorithm consists of both feature detection and representation aspects. It is designed to produce features akin to those produced by a SIFT descriptor on Hessian-Laplace interest points, but using efficient approximations. Reported results indicate that SURF provides a significant speed-up while matching or improving performance.

Other descriptors which have been proposed include Gabor filter banks, image moments, and others. A study by Mikolajczyk and Schmid compares several feature descriptors, and shows that SIFT-like descriptors tend to outperform the others in many situations \citep{mikolajczyk_performance_2005}. The descriptors that were evaluated, however, lack color information. This is in contrast to the biomimetic vision community which typically includes a color-opponency aspect to feature representations. There is evidence that including color information in feature detection and description may improve BoF \ir{} performance \citep{jiang_towards_2007}. A recent paper by van de Sande et al. presents an evaluation of color feature descriptors. Reported results indicate a combination of color descriptors outperforms SIFT on an image classification task and that, of the color descriptors, OpponentSIFT is most generally useful \citep{sande_evaluating_2010}.

Finally, there has been investigation on learning discriminative features for a given task or data set, instead of using an a-priori selected representation. Efforts include a method for unsupervised learning of discriminative feature subsets and their detection parameters  \citep{karlinsky_unsupervised_2009}, and a modular decomposition of feature descriptors and a method for learning the best composition \citep{winder_learning_2007,winder_pickingbest_2009}. Winder demonstrates that many common feature descriptors, such as SIFT, can be generated this way, yet there are others that can be learned that perform better under certain measures.

\section{Quantization and Distance Measures}
\label{section:quantization}

Vector Quantization (Clustering) is used to build the visual vocabulary in \bof{} algorithms. Nearest-neighbor assignments are used not only in the clustering of features but also in the comparison of term vectors for similarity ranking or classification. Thus, it is important to understand how quantization issues, and the related issues involving measuring distances in feature and term vector space, affect \bof{} based applications.

There are a great many clustering/vector quantization algorithms, and this report does not attempt to enumerate them. Many BoF implementations are described as using K-means \citep{sivic_video_2003,lazebnik_beyond_2006,jiang_towards_2007}, or an approximation thereof for large vocabularies \citep{nister_scalable_2006,philbin_object_2007}. Given any clustering method, there will be points that are equally close to more than one centroid. These points lie near a Voronoi boundary between clusters and create ambiguity when assigning features to terms. With K-means and similar clustering methods, the choice of initial centroid positions affects the resultant vocabulary. When dealing with relatively small vocabularies, one can run K-means multiple times and select the best performing vocabulary during a validation step. This becomes impractical for very large data sets. When determining the distance between two features, as required by clustering and term assignment, common choices are the Manhattan ($L_1$), Euclidean ($L_2$), or Mahalanobis distances. A distance measure is also needed in term vector space for measuring the similarity between two images for classification or retrieval applications. Euclidean and Manhattan distances over sparse term vectors can be computed efficiently using inverted indexes (see \ref{sect:scale}), and are thus popular choices. However, the relative importance of some visual words leads to the desire to weight term vectors during the distance computation. The following sub-sections provide more details on these and other issues.

\subsection{Term Weights}

One of the earliest strategies for handling quantization issues at a gross level is to assign weights to the terms in the term vector. This can be viewed as a mitigation strategy for quantization issues that occur when the descriptors are distributed in such a way that simple clustering mechanisms over-represent some descriptors and under-represent others. With term weights, one can penalize terms found to be too common to be discriminative and emphasize those that are more unique. This is the motivation behind the popular Term Frequency-Inverse Document Frequency (TF-IDF) weighting scheme used in text retrieval \citep{salton_introduction_1983}.

TF-IDF is defined as: $tf_i \cdot log(\frac{N}{N_i})$, where $tf_i$ is the term frequency of the i'th word, $N$ is the number of documents (images) in the database, and $N_i$ is the number of documents in the database containing the i'th word. The log term is called the inverse document frequency and it serves to penalize the weights of common terms.

Term vectors can also be represented as binary strings. A $1$ is assigned for any term that appears in the image, $0$ otherwise. This might humorously be called ``anti term weighing,'' as it has the effect of making the terms have equal weight no matter how often they occur in a particular image or in the corpus as a whole. Distribution issues are thus handled by simply erasing the frequency information from the term vector so a few oversampled terms can not dominate the distance computations between term vectors. Binary representations have the benefit of speed and compactness.

BoF \ir{} implementations typically use TF-IDF weights, due to evidence that this method is superior to binary and term frequency representations \citep{jiang_towards_2007,sivic_video_2003}. When using very large vocabularies, the term vectors become extremely sparse, and the term counts (prior to any normalization) are mostly zeros or ones. In this case, binary representations tend to perform similarly to term frequencies \citep{jiang_towards_2007}.

\subsection{Soft Assignment}

A given feature may be nearly the same distance from two cluster centers, but with a typical ``hard assignment'' method, only the slightly nearer neighbor is selected to represent that feature in the term vector. Thus, the ambiguous features that lie near Voronoi boundaries are not well-represented by the visual vocabulary. To address this problem, researchers have explored multiple assignments and soft weighting strategies.

Multiple assignment is where a single feature is matched to $k$ nearest terms in the vocabulary. Soft weights are similar, but the $k$ nearest terms are multiplied by a scaling function such that the nearest term gets more weight than the $k$'th nearest term. These strategies are designed to mitigate the negative impact when a large number of features in an image sit near a Voronoi boundary of two or more clusters.

Jegou et al. show that multiple assignment causes a modest increase in retrieval accuracy \citep{jegou_contextual_2007}. The cost of the improved accuracy is higher search time, due in part to the impact on term vector sparsity. The authors report that a $k=3$ multiple assignment implementation requires seven times the number of multiplications of simple assignment.

Soft weights have been explored by \citet{jiang_towards_2007} and \citet{philbin_lost_2008}. In Jiang's work, the soft weights are computed as shown below. The computed weight for term $n$ in the term vector $w$, denoted $w_n$ is defined as:

\begin{equation}
	w_n = \sum_{i=1}^{k} \sum_{j=1}^{M_i} \frac{1}{2^{i-1}} sim(j,n)
\label{eqn:JiangSoftW}
\end{equation} 
where $k$ is the number of neighbors to use in the soft weighting strategy, $M_i$ is the number of features in the image whose i'th nearest neighbor is term $n$, and $sim(j,n)$ is the similarity measure between feature $j$ and term $n$. Jiang et al. suggest that $k$=4 works well. In experimental evaluations, over a variety of vocabulary sizes, Jiang's soft weighting strategy bests binary, term frequency, and TF-IDF schemes, with one marginal exception.

Philbin et al. propose an approach that scales the term weight according to the distance from the feature to the cluster center \citep{philbin_lost_2008}. The weights are assigned proportionally to a Gaussian decay on the distance, $exp(-\frac{d^2}{2\sigma^2})$, where $d$ is the distance to the cluster center, and $\sigma$ is the spatial scale, selected such that a relatively small number of neighbors will be given significant weight. A problem with this continuous formulation is that all terms get a non-zero weight, so clipping very small values is prudent. Philbin et al. only compute the soft weights to a pre-determined number of nearest neighbors, which was three in their evaluations. Even with clipping, soft weights decrease the sparsity of the term vectors, and thus increase the index size and query retrieval times. After generating term vectors using the soft weighting strategy, Philbin et al. perform an $L_1$ normalization so that the resulting vector looks like a term frequency vector. TF-IDF is then applied, ignoring soft assignment issues -- i.e., the IDF is computed as if the input vector were created by the normal hard-assignment process. Evaluation results show a strong improvement in query accuracy.

This result is consistent with the earlier observations by Jegou et al. and Jiang et al. that multiple assignment/soft weighting improves retrieval accuracy by mitigating some of the quantization errors for borderline features. In the body of literature surveyed by this report, no direct comparison between these three methods has been performed.

\subsection{Non-uniform Distributions}

Jurie and Triggs show that the distribution of cluster centers is highly non-uniform for dense sampling approaches. When using k-means clustering, high-frequency terms dominate the quantization process, yet these common terms are less discriminative than the medium-frequency terms that can get lost in quantization. Instead, they propose an online, fixed-radius clustering method that they demonstrate produces better codebooks \citep{jurie_creating_2005}. Jegou et al. discuss the ``burstiness'' of visual elements, meaning that a visual word is more likely to appear in an image if it has appeared once before. Thus visual words are not independent samples of the image. Various weighting functions and strategies are proposed and evaluated \citep{jegou_burstiness_2009}.

Similar to distribution issues within the feature space used to construct the vocabulary, term vectors can be non-uniformly distributed in the gallery set. Weighting strategies can be used to compensate for non-uniform term vector distributions, as discussed above, or a distance measure can be created that scales with local distributions in an attempt to regularize the space.

\begin{figure}[ht]
  \centering
  \includegraphics[width=2.5in]{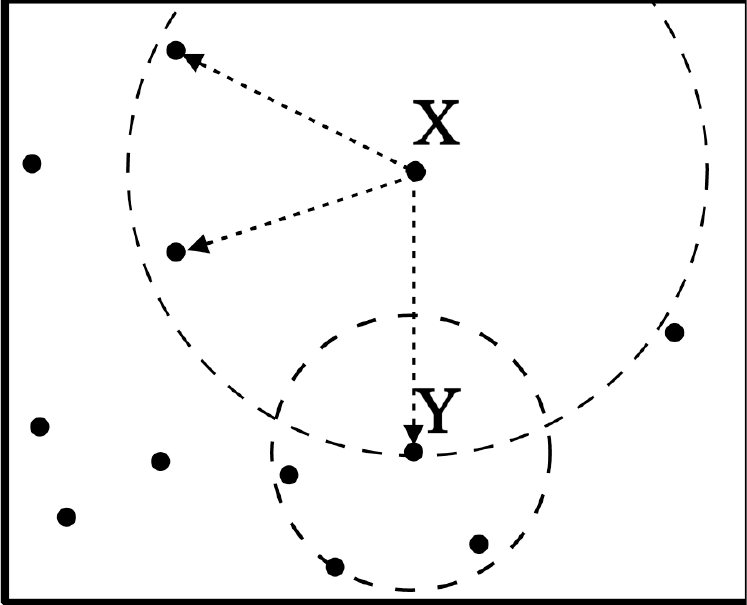}
	\caption{Illustration showing nearest neighbors ``nonreversibility.'' Point Y is a 3-nearest neighbor of X, but the converse is not true. The Contextual Dissimilarity Measure attempts to address this issue.}
	\label{fig:nn_issue}
\end{figure}

The latter approach is explored by the Contextual Dissimilarity Measure (CDM) \citep{jegou_contextual_2007}. Jegou et al. point out the non-symmetry in a k-nearest-neighbors computation, which is central both to the vocabulary generation in the feature space and also to computing similarity scores over the term vector space. Consider that some point $x$ may be a neighbor of $y$, but the converse may not be true, as seen in Figure \ref{fig:nn_issue}. They call this effect ``neighborhood nonreversibility,'' and implicate it as part of a problem in BoF-based \ir{} that causes some images to be selected too often in query results while others are never selected at all. At a high level, CDM regularizes the term vector space by penalizing the distance measure for points in a local neighborhood that cause nonreversibility.

A discussion of the details of CDM is beyond the scope of this survey, but Jegou et al. show the CDM computation results in a single scalar distance update term for each term vector in the gallery. Computation of the exact CDM values, performed offline while indexing the images, is of quadratic complexity in the number of images. Fortunately, the authors show that approximate methods work well, so the quadratic complexity is mitigated in practice. The distance between a query term vector and a gallery term vector is simply the distance multiplied by the CDM update term for that gallery image. Any distance function appropriate to the term vector space can be used. Given a distance function represented as $d(q,v_j)$, with $q$ as the query term vector, $v_j$ as the term vector for the j'th image in the gallery, and $\delta_j$ as the distance update term for the j'th term vector, the Contextual Dissimilarity Measure is simply:

\begin{equation}
 CDM(q,v_j) = \delta_j * d(q,v_j)
\label{eqn:CDM}
\end{equation}

Evaluation results of applying the CDM regularization to an otherwise standard BoF \ir{} implementation shows significantly improved accuracy over contemporary methods. As a side note, their evaluations also show that the Manhattan ($L_1$) distance works better than the Euclidean ($L_2$), whether or not CDM is applied.

\subsection{Intracluster Distances}

There is a trade-off involved in choosing the granularity of the quantization, where finer-grained clustering leads to potentially more discriminative information being preserved at the cost of increased storage and computational requirements. A possible compromise is to use a courser-grained quantization, but compensate for the lack of discrimination by employing an efficient method for incorporating intracluster distances to weight terms. To this end, Jegou et al. developed a technique called Hamming Embedding, which efficiently represents how far a feature lies from the cluster center and thus how much weight to assign to that term for the detected feature \citep{jegou_hamming_2008}. 

\section{Image Classification using \bof}
\label{section:IC}
In the preceding sections, we have discussed BoF image representations somewhat independently of the task. In this section and the next, we explore the two most common BoF applications in the literature, \ic{} and \ir{}. We start with \ic{}.

\subsection{Problem Definition}
\label{sect:ICPD}
We use the term \textit{\ic{}} to describe any supervised classification of images as a whole, as opposed to classification based more directly on the specific contents (objects) present in the image. BoF based \ic{} is the process of representing a training set of images as term vectors and training a classifier over this representation. A probe image can be encoded with the same dictionary and given to the classifier to be assigned a label.

Similar to the \ic{} task, people discuss BoF-based object detection, object recognition, and scene classification. When the data set has essentially one object class per image, then the difference between \ic{} and object detection is blurred. This is the case, for example, in the popular CalTech101 data (see \citep{fei-fei_learning_2004}). Due to the discarded spatial information, BoF representations are not well-suited to object detection and localization, but there are examples in the literature of doing so in an approximate fashion (\citep{sivic_discovering_2005}, e.g.). Scene classification is essentially synonymous with \ic{}. That is, the entire image is classified with no attempt to detect or localize specific objects it may contain.

\subsection{Related Methods}

To help put the BoF \ic{} literature in context, we first investigate a set of related approaches. The first is a brief discussion of the similarity between BoF \ic{} and texture classification. As with \bof{}, texture classification methods commonly sample features from the image (a texture image), quantize those features, and build representative histograms, which are then used in a classification task (see \citet{leung_representing_2001,liu_spectral_2002}, among many others). In the texture recognition literature, ``textons'' refer to representative image patches and are analogous to visual words in the \bof{} paradigm. Zhang et al. performed a study of local features and classification kernels for Support Vector Machine-based texture and image classification \citep{zhang_local_2005}, which indirectly illustrates how similar \bof{} \ic{} is to the earlier body of work on texture representation and classification.

A second body of related work is object detection using a part-based model. Unlike the orderless \bof, a part-based model (also called a ``constellation'') learns a deformable arrangement of features that represent an object class. Similar to BoF, part-based models often start with feature detection and extraction stages. Diverging from a BoF approach, part-models typically employ a maximum likelihood estimation technique to determine which of several previously learned models best explain the detected arrangement of features. Part-based methods have shown strong robustness in the face of background clutter, partial occlusions, and variances in the object's appearance \citep{burl_probabilistic_1998,crandall_weakly_2006,fergus_object_2003,leordeanu_beyond_2007,sudderth_learning_2005}. A key weakness of part-based models is computational complexity. There is a combinatorial search required to determine which subset of features matches a given model. To reduce computation time, often only a small number of features are extracted from the image (one or two orders of magnitude smaller than with BoF methods), thus making the algorithms more sensitive to feature detection errors.

Another related technique is the use of a gist descriptor for scene classification, such as the Holistic Spatial Envelope \citep{oliva_modelingshape_2001}. Gist-based approaches attempt to create a low-dimensional signature of an image that is computationally cheap and contains enough information for gross categorization. Both gist and BoF representations attempt to reduce an image to a low dimensional representation, but while BoF is a histogram of quantized local features, gist is a single global descriptor of the entire image. Gist representations are thus smaller and faster to compute, but with a commensurate loss of discriminatory information. The Holistic Spatial Envelope is akin to using a single 512-dimensional SIFT descriptor to represent the entire image, as described in \citep{torralba_small_2008}.

Visual localization is the problem of determining approximate location based only on visual information. It is a related task to \ic{}, because input images must be classified as being part of a known location. Researchers have been using SIFT features as natural landmarks for many years \citep{se_mobile_2002,goedem_omnidirectional_2005,spexard_biron_2006}. Recently, researchers have applied BoF representations directly to the task \citep{fraundorfer_topological_2007,kang_image_2009,kazeka_visual_2010,naroditsky_videotrek:vision_2009}.

Wrapping up our discussion of works relating to BoF \ic{} are biologically inspired methods for scene classification. A well-known example is scene classification using the Neuromorphic Vision Toolkit \citep{siagian_rapid_2007}. Siagian and Itti detect local color, intensity, and gradient orientation features using non-maxima suppression over a scale space, but in contrast to BoF representations, they impose a spatial structure by dividing each image into a 4x4 grid, over which features are aggregated. More recently, Song and Tao presented a ``biologically inspired feature manifold'' representation of an image \citep{song_biologically_2010}. With the assumption that the features are sampled from a low dimensional manifold embedded in a high dimensional space, the authors propose a manifold-based projection to reduce dimensionality while preserving discriminative information. They show SVM-based scene classification on this representation yields much faster and more accurate results than that of Siagian and Itti \citep{song_biologically_2010}.

\subsection{The BoF Approach}
This section provides an overview of key papers on BoF \ic{}, in roughly chronological order to show how the technique has evolved over the past several years.

Csurka et al. present an early application of the \bof{} representation for \ic{} \citep{csurka_visual_2004}. Their front-end process closely follows the one outlined in Section \ref{section:BOF}. They employ the Harris-Affine keypoint detector and the 128-dimension SIFT descriptor. They generate several vocabularies using k-means clustering, varying the value of $k$. They compare the use of Naive Bayes and linear Support Vector Machine (SVM) classifiers to learn a model which can predict the image class from its BoF representation. Experimental results show that SVM strongly outperforms Naive Bayes, and that larger vocabulary sizes (as measured by $k$, the number of cluster centers) perform better, within the tested range of 100-2500. To account for the random starting positions of k-means, ten vocabularies for each choice of $k$ were tested, with the best results reported.

Jurie and Triggs investigate BoF-based \ic{}, with an emphasis on sampling strategies and clustering methods for creating vocabularies \citep{jurie_creating_2005}. Features are extracted by densely sampling grayscale variance-normalized 11x11 patches over a multiscale grid. Vocabulary creation uses a novel Mean-Shift-like clustering method that is purported to better account for the highly non-uniform densities in feature space. Naive Bayes and linear SVM classification results are reported, using 10-fold cross validation, showing SVM outperforming Naive Bayes in all tests. Codebooks were generated using 2500 cluster centers. The authors compare their method (dense pyramid sampling and novel clusterer) with two others: dense sampling with k-means clustering, and a Difference of Gaussian (DoG) keypoint detector with k-means clustering. Their method performs best; the keypoint-based method performs worst. They conclude that sparse, keypoint-based representations fare poorly due to a loss of discriminative power, and that with dense sampling, k-means produces poor codebooks compared to one that enforces fixed-radius clusters.

Zhang et al. perform a large-scale quantitative evaluation of BoF representations for both texture recognition and \ic{} \citep{zhang_local_2005}. This evaluation looks at different feature detectors, different region descriptors, and SVM classifier kernels. While there was no single best method for all tests, the authors recommend a mix of detectors and features with complementary types of information. Further, the authors point out that using local features with the highest level of invariance does not yield the best results.

Continuing along the same lines as their earlier work, Nowak, Jurie, and Triggs further explore sampling strategies and other factors impacting BoF-based image classification \citep{nowak_sampling_2006}. In contrast to their earlier work, this implementation uses the SIFT descriptor, showing that it is superior to normalized pixel intensities. This work reinforces the fact that random dense sampling, assuming a high-enough number of keypoints per image, outperforms keypoint detectors for \ic{} with SVMs. They present results on the 2005 PASCAL Visual Object Classification challenge (see \citet{everingham_2005_2006}), showing superior accuracy in all categories when compared to the best individual results for each category. As such, it set a new high-water mark for image classification while using simpler methods than its contemporaries.

Fei Fei and Perona present a BoF \ic{} approach based on a generative Bayesian hierarchical model \citep{feifei_bayesian_2005}. Their model includes latent topics, or ``themes,'' that are considered hidden variables. An algorithm based on Latent Dirichlet Allocation is applied to learn the model given weak supervision in the form of image class labels. While the accuracy of this approach may not compare to more recent SVM-based classification methods, it nevertheless is attractive for the ability to learn intermediate-level themes present in an image based on a \bof{} representation. Their results show that the SIFT descriptor is superior to normalized 11x11 pixel intensities for their method. They also show that a dense grid-based sampling technique outperformed the two keypoint operators they evaluated, with the observation that the grid sampling technique generated the most feature points per image.

Sivic et al. propose using Probabilistic Latent Semantic Analysis (pLSA) with a \bof{} representation for \ic{} and even object localization \citep{sivic_discovering_2005}. Conceptually similar in some respects to Fei Fei and Perona's contemporary work (discussed above), Sivic et al. employ a probabilistic model that discovers hidden (latent) topics within the images based on the observed BoF representation. In this case, the object classes are considered the latent topics, such that an image containing a mixture of objects from different classes would be modeled as a mixture of latent topics. Feature detection is performed using both MSER and Harris-Affine keypoint operators. Features are represented by SIFT descriptors. As with the other techniques presented in this section, weak supervision is provided in the form of image labels. A notable contribution includes a method for localizing the objects (via rough segmentation) by selecting groups of maximally-likely features for a given topic. They also introduce visual word ``doublets,'' learned from spatially co-occurring features. Results indicate that using doublets helps increase the localization accuracy.

Grauman and Darrell present the Pyramid Match Kernel for use with SVM-based image classification for BoF representations \citep{grauman_pyramid_2005}. In the Pyramid Match Kernel, each feature set is mapped to a multi-resolution histogram. Histogram pyramids are matched with a weighted histogram intersection function, where courser bins have less weight than finer. This kernel approximates the optimal correspondence matching between sets of unequal cardinality, and is computationally faster than many comparable kernels. The Pyramid Match Kernel is a Mercer kernel (positive semi-definite), and thus SVM convergence is assured.

Inspired by the Pyramid Match Kernel, Lazebnik et al. introduced the Spatial Pyramid Matching technique for \ic{} \citep{lazebnik_beyond_2006}. The key difference between the two approaches is that the Pyramid Match Kernel uses a multi-resolution histogram while Spatial Pyramid Matching uses a fixed sampling resolution but changes the size of the regions over which the histograms are formed. The Spatial Pyramid Match kernel is represented as the sum of pyramid match kernels over the quantized vocabulary. The Spatial Pyramid Matching technique no longer represents an image as an orderless collection of features, thus it is not a true \bof{} approach. However, it does reduce to a BoF approach when only a single pyramid level is used. Grauman and Darrell's Pyramid Match Kernel approach, on the other hand, maintains the orderless feature representation, but does not use a visual vocabulary. Lazebnik et al.'s Spatial Pyramid Matching uses SIFT descriptors extracted from a dense grid with spacing of 8 pixels. K-means clustering is used for constructing the vocabulary, which consists of either 200 or 400 clusters.

In addition to presenting the Spatial Pyramid Matching kernel, Lazebnik et al. provide evidence that latent factor analysis, such as that in \citep{feifei_bayesian_2005,sivic_discovering_2005}, hinders classification accuracy. The comparison is done using a single-level pyramid that is equivalent to a standard BoF representation with a 200-term vocabulary, using the same data set and protocol as \citep{feifei_bayesian_2005}. They outperform Fei Fei and Perona's method by about 7\%. When Lazebnik et al. apply pLSA to their model, their results drop back to be comparable with Fei Fei and Perona.

Jiang, Ngo, and Yang evaluate several factors that impact BoF \ic{} using SVMs \citep{jiang_towards_2007}, most notably the choice of feature detector, the term vector weighting scheme, and the SVM kernel. SIFT descriptors are used in all evaluations. Using term frequency, 1,000 visual words, and a $\chi^2$ RBF kernel, Jiang et al. show that the DoG interest point operator outperforms other popular methods on the PASCAL-2005 data set. Note that the experiment controlled the average number of keypoints per image to between 750 and 925 for all detectors. Jiang et al. compare term weighting strategies as well, evaluating binary, term frequency, TF-IDF, and soft weighting. For all but the smallest vocabulary size used in the evaluation on the PASCAL-2005 data, soft weighting proved superior. SVM Kernels that were evaluated by the authors include Linear, Histogram Intersection (HI), Gaussian RBF, Laplacian RBF, Sub-linear RBF, and $\chi^2$ RBF. On the PASCAL-2005 data set, the best mean equal error rates occurred for the latter three of the six kernels. The authors subsequently recommend the $\chi^2$ RBF and Laplacian RBF kernels. Combining their recommended selections of DoG feature detector, soft weighting scheme, and $\chi^2$ RBF kernel, they show results comparable to the then-current state-of-the-art performance on the PASCAL-2005 data set. They note that their performance is close to that of Nowak et al. \citep{nowak_sampling_2006} (the best reported result at the time), but that they use far fewer features per image compared to Nowak's dense sampling strategy.

\section{Image Retrieval using \bof}
\label{section:IR}

\subsection{Problem Definition}
Another popular application of the \bof{} image representation is \ir{}. We define \ir{} as the task of finding the most similar images in a gallery to a given query image. The gallery is a database of images, often from video. We differentiate this from Content-Based Image Retrieval (CBIR) approaches that try to find those images which contain a set of specific objects or other concepts, like ``sky.'' In \ir{} as we define it, the process is query-by-example using an entire image or image subregion as the example. A \bof{} based \ir{} algorithm returns results based on a similarity score (distance) between the query image term vector and the term vectors of the gallery images, ranked accordingly. Note that the BoF \ir{} approach requires no supervised training. This is an important strength of the approach, allowing the efficient indexing of large image sets with no ground truth training data.

In contrast, a common CBIR approach is to create a large number of specific object/concept detectors and to index the gallery set based on the detection results for each. A query can be specified as various combinations of these objects/concepts, perhaps weighing some objects as being more important to the user than others, and results returned based on the previously indexed detection results. The disadvantage of this approach is the necessity of creating a seemingly endless number of object detectors, each requiring ground truth labels, parameter selection, validation, and so on.

The similarity-based \ir{} approach requires the user to present one or more sample images of what he is searching for. The flexibility and efficiency of BoF \ir{} comes with the trade-off that it puts more effort on the user to generate a query and lacks the semantics required to support text-based queries.

\subsection{Related Methods}

A recent survey of CBIR algorithms was conducted by \citet{datta_image_2008}. Datta uses the term CBIR more generally than we have defined it here. His survey includes a wide variety of image/video query technologies. As our focus is the BoF representation, we explore this \ir{} method in more detail and focus. We refer the interested reader to Datta's survey for a comprehensive look at CBIR and the similarities between various approaches.

Gist descriptors have also been applied for \ir{}. Gist was discussed in Section \ref{sect:ICPD} in the context of \ic{}, but just as the BoF representation has been applied to multiple applications, so has gist. Torralba et al. developed compact gist-based representations used to retrieve results from a gallery of 12 million icon-sized images \citep{torralba_small_2008}. In a variant of the \ir{} task applied for image/video copy detection, Douze et al. compared the relative utility of gist and BoF representations, with the expected trade-off of speed vs. accuracy \citep{douze_evaluation_2009}.

\subsection{The BoF Approach}

A seminal work defining the \bof{} \ir{} approach is the ``Video Google'' paper of \citet{sivic_video_2003}. Video Google uses both MSER and Harris-Affine keypoint detectors to detect features, which are represented by SIFT descriptors. The vocabulary is built using k-means clustering. Nearest Neighbor term assignment and the Euclidean distance on TF-IDF weighted term vectors are used for similarity scoring. Additionally, Video Google employs a ``spatial consistency'' step that attempts to validate the search results by aligning subsets of interest points with an approximate affine transformation.

While Video Google demonstrated the effectiveness of the BoF representation for \ir{}, it was far from being ``Google-like'' in scale. Since then, researchers have extended this basic approach to (1) deal with much larger scale data sets, (2) improve the initial query results, and (3) improve results ranking using post-query rank adjustment. Of these three major areas, the improvements to initial query results have stemmed from improving feature detection and representation (see Section \ref{section:FDE}) and addressing quantization issues (see Section \ref{section:quantization}). Improvements relating to (1) and (3) are discussed below. Following that, we discuss the challenge of trying to generate a general visual vocabulary which could be used to index any image set.

\subsubsection{Scalability}
\label{sect:scale}
To perform Internet-scale \ir{}, \bof{} methods require efficient indexing strategies and the ability to handle large vocabularies. In Video Google, the number of images in the database was approximately 4,000, and the visual vocabulary was on the order of 10,000 terms. An inverted file system structure was used to make queries efficient, resulting in query response time of about 0.1 seconds over 4,000 images.

Essentially, an inverted file system (or inverted index) method for indexing the gallery term vectors is one where each term records the images in which it appears, along with the term frequency for each image. The motivation for the inverted index is to take advantage of the sparseness of the term vectors. If each image yields approximately 2,000 interest points and the vocabulary size is 10K terms, then the resulting term vector (assuming no multiple/soft assignments) will have at most 20\% of its entries as non-zero. In a larger system with a vocabulary of 100K to 1M terms, the term vectors will be even sparser.

A simple approach to ranking query results is to compute the distance between the query vector and the term vectors of each gallery image, requiring an $O(N)$ computation in term of the number of gallery images. The majority of those images are likely to have very few terms in common due to the sparsity of the vectors, however. Instead, one can look at each non-zero term in the query vector and get the list of images in which that term appears via the inverted index. Compiling this list for all non-zero terms will result in a subset of the gallery that contains partial matches. The normalized $L_p$ distance to each image discovered in the inverted index can be computed incrementally as the inverted index is traversed, such as demonstrated in \citep{nister_scalable_2006}. While the worst case complexity remains $O(N)$, there is a huge speed improvement for the average case. Inverted indexes are used in all contemporary BoF-based \ir{} methods that were surveyed in this report.

Additionally, the use of a \textit{stop-words} list, a technique from text retrieval that eliminates the most common and least discriminative words (``a,'' ``the,'' etc.), can also be used in \ir{}. In this case, one simply eliminates any term from the vocabulary that appears in too many images. Stop-words were noted as being helpful in \citep{sivic_video_2003,nister_scalable_2006}, and clearly are helpful in avoiding the traversal of a long list of documents in an inverted file system.

However, even with the use of an inverted index and stop-words, more needs to be done to achieve acceptable performance in very large data sets. Nister and Stewenius made a scalability breakthrough by applying a hierarchical clustering method to represent the vocabulary as a tree \citep{nister_scalable_2006}. Compared to the Video Google approach, Nister and Stewenius demonstrated that their vocabulary tree approach allows for much larger vocabularies, with results reported on vocabulary trees having 1M leaf nodes, and an anecdotal mention of up to 16M leaf nodes. They use an inverted index listing at each node in the vocabulary tree and show how to efficiently compute the term vector similarity score over their tree structure. Their large-scale data set was constructed by embedding 6376 ground-truth similar images into a large corpus generated from every frame of seven full length movies, yielding over 1 million images. They report that queries on this data set run in RAM on an 8GB machine and take approximately 1 second each. Index creation took 2.5 days, which was dominated by feature extraction time. A benefit of the vocabulary tree approach is that, in principle, one can build the tree online and incrementally, which may make it suitable for tasks such as visual location awareness in mobile robots, where the robot incrementally builds its vocabulary as it explores its surroundings.

Philbin et al. use a different approach by employing an approximate k-means clusterer during vocabulary creation \citep{philbin_object_2007}. Their approximate k-means (AKM) algorithm uses a forest of randomized k-d trees. Lowe describes the use of a k-d tree to approximate nearest neighbor computations of SIFT descriptors \citep{lowe_distinctive_2004}, and others have further developed the concept for increased speed and accuracy \citep{silpa-anan_localisation_2004,silpa-anan_optimised_2008,moosmann_randomized_2006,moosmann_fast_2007,muja_fast_2008}. Philbin et al. use an implementation provided by Lowe, which reduces the complexity of a single k-means iteration from $O(Nk)$ to $O(N\log(k))$, which is the same complexity as Nister's hierarchical k-means (HKM) approach. Philbin et al. compare AKM with HKM, and claim equivalent speed but significantly greater accuracy. They conclude that AKM suffers less from quantization errors than HKM, and thus is superior.

Other scalable indexing methods have been explored, including Locality Sensitive Hashing (LSH) \citep{ke_efficient_2004,andoni_near-optimal_2006}, dimensionality reduction of BoF vectors \citep{chum_near_2008,jegou_packing_2009}, and Product Quantization \citep{jegou_aggregating_2010}.

Another issue for massive image databases is the potential for parallelization of the query process. Philbin et al. note that their approach which uses a forest of k-d trees could be distributed to multiple servers. No significant work has been demonstrated to parallelize BoF based \ir{} to our knowledge, however.

\subsubsection{Post Query Processing}
\label{sect:ranking} 

There have been several techniques put forth in the literature for improving the performance of a \bof{} \ir{} process by post-query processing of the result set. As mentioned previously, Sivic et al. employ a spatial consistency re-ranking process that alters the initial query results based on how well an estimated affine homography maps the feature points between the query and gallery image \citep{sivic_video_2003,philbin_object_2007,philbin_lost_2008}. While this method does improve the results, it adds significant additional computation and may not be feasible for massive data sets.

Another way to improve query results is through rank aggregation \citep{jegou_accurate_2010}. In rank aggregation, the query is performed multiple times using a separate vocabulary (index) for each query. The results are aggregated by, for example, taking the mean rank of the results. This increases the query time and index size. The separate queries can be processed in parallel to mitigate the time penalty, but there is still the cost of aggregation. From a storage perspective, multiple indexes increase the space requirements.

A third post-query mechanism is query expansion, as presented by Chum, et al. in \citep{chum_total_2007}. With query expansion, top-ranked results from the initial query are re-submitted as additional queries in an attempt to increase recall at a given precision. Chum et al. use spatial consistency to help prevent false positives from being used in the query expansion. Their results show this is a critical step, as query expansion without spatial verification performs worse than not using query expansion at all. The authors compare several different query expansion strategies, the best of which increases the mean Average Precision retrieval score on the Oxford data set by 20 percentage points over baseline. Information about the mean Average Precision metric and the Oxford data set can be found in Section \ref{section:eval}.

\subsubsection{Generalization}
Ideally, one might like to have a universal mechanism for \ir{} that does not require training on a specific data set. In most of the surveyed BoF \ir{} approaches, the vocabulary is trained on a subset of the gallery images. While this subset and the gallery set used for testing may be separate, there is still the concern that the vocabulary may be optimized for use on the data set as a whole. How representative of the gallery data does the vocabulary training data need to be for effective \ir{}? If it must be somewhat representative, then how does one train a vocabulary that will work well for billions of web images?

While there is no conclusive response to those questions, there are some encouraging results in the surveyed literature. Nowak et al. show that a codebook created from randomly generated SIFT vectors can be effective for BoF-based \ic{} \citep{nowak_sampling_2006}. Their study compared three codebooks evaluated over two databases: the Graz object set \citep{opelt_weak_2004} and KTH texture image set \citep{hayman_significance_2004}. The three codebooks used are: the random SIFT codebook, a codebook created from a subset of the Graz images, and a codebook created from a subset of the KTH images. Unsurprisingly, the KTH codebook yields the best classification accuracy on the KTH images and the Graz codebook is best on the Graz images. More interesting, perhaps, is that the random codebook has approximately the same error rates on both sets, increasing the error rate by 5-10 percentage points over the set-specific codebook. Also of note is that the Graz codebook scores second-best on the KTH data and, similarly, the KTH codebook is second-best on the Graz data. This indicates that a codebook generated on a ``real'' image space performs better than a purely random codebook from the same feature space. While random codebooks have not been evaluated for \ir{}, it is reasonable to believe the effects reported for \ic{} might carryover.

Supporting the notion that it may be possible to create a generic, or universal, codebook is anecdotal results reported by the Video Google project \citep{sivic_video_2003}. They report \ir{} was effective even when using a codebook generated from \textit{Run Lola, Run} to index and retrieve images from the movie \textit{Groundhog Day}.

\section{Evaluation}
\label{section:eval}
This section provides an overview of the common methods for evaluating the performance of \bof{} systems. A summary is presented of the quantitative metrics used to evaluate performance and the common benchmark data sets. Finally, this section concludes with the challenges in comparative evaluation and what could be done to improve the situation.

\subsection{Performance Metrics}
One of the most common performance measures for \ir{} is the precision-recall curve. Precision is defined as the ratio of true positives returned by a query over the total number of results returned. Recall is the ratio of true positives returned by a query over the total number of true matches possible in the gallery. It has been said that state-of-the-art BoF-based \ir{} methods typically have high precision at low recall \citep{chum_total_2007}, meaning that the top results are highly relevant to the query, but many other relevant images are not returned. From the precision-recall curve, one can define the Average Precision (AP) as the area under the curve. A perfect precision-recall curve would have a perfect precision (1.0) at all recall levels, thus the maximum AP is 1.0. A variant of AP used by Philbin et al. \citep{philbin_object_2007} is the mean Average Precision (mAP), which is the AP averaged over 5 different query regions from the same landmark on the Oxford data set. An Equal Error Rate (EER) for precision-recall curves is also used by some, and is the value on the curve where precision equals recall.

Sponsored by the U.S. National Institute of Standards and Technology (NIST), the TRECVID community uses Inferred Average Precision (InfAP). InfAP is an approximation to AP used when the dataset is too large or dynamic to allow for complete relevancy judgments on ranked results. InfAP treats the AP measure as the expectation over a random experiment, and is described in \citep{yilmaz_estimating_2006}. 

Another performance measure proposed for \ir{} is the Average Normalized Rank (ANR) \citep{sivic_video_2003}. Given a known number of images that should ideally be retrieved for a given query, ANR computes a measure of the actual ranking versus the ideal, normalized by both the number of relevant images and also the size of the gallery. ANR is computed as follows, where $N$ is the number of gallery images, $N_r$ is the number of relevant images, and $R_i$ is the actual rank of the i'th relevant image:

\begin{equation}
ANR = \frac{1}{N\;N_r} ( (\sum_{i=1}^{N_r} R_i) - \frac{N_r(N_r+1)}{2} )
\label{eqn:ANR}
\end{equation}

To help understand Equation \ref{eqn:ANR}, consider an example of five relevant images returned from a database of 1000 with the following ranks: \{3,4,8,100,400\}. Note that the order of the five images does not matter, one of them is ranked 3rd, one is 4th, one is 8th, etc. The perfect set of rankings would be: \{1,2,3,4,5\}. The term $\frac{N_r(N_r+1)}{2}$ simply sums the values 1 to $N_r$, representing the perfect score. In this example, ANR = $\frac{1}{1000*5} ( (3+4+8+100+400) - 15) = \frac{500}{5000} = 0.10$. The interpretation is that the average ranking of a relevant image in this query was in the top 10\%. The smaller the number, the better. A perfect ANR score is zero.

One final performance measure discussed in the literature is the Nister-Stewenius (NS) score, developed for use with the NS data set. The NS data set has 4 images each of 2,550 objects. The NS score indicates the average number of the four images that are returned in the results for a given query. Note that in this setup, each of the four images is used as the query, and is also still part of the gallery. One expects the identical image will be returned in the query, thus while the minimum NS score is technically zero, only a very poor algorithm would fail to find an exact match. So the NS score effectively ranges from 1 to 4, where 4 is a perfect result of all four images being returned when any one of them is given as the query.

\subsection{Comparative Evaluation}
While comparing \ic{} approaches is relatively straight-forward given a common benchmark data set, there are challenges in trying to compare BoF \ir{} methods. The key challenge for comparative evaluation is the lack of consensus on performance measures, data sets, and evaluation protocols. NIST has been running the TRECVID challenge for video information retrieval for nearly a decade. Perhaps the reason why leading BoF \ir{} systems do not present TRECVID results is that the problem is slightly different. With the TRECVID information retrieval task, one must find those video frames that contain certain pre-defined concepts. With Video Google and its more contemporary progeny, the task is to find those images that best match a given image or image region. The TRECVID challenge is more of an \ic{} or object detection challenge in a large scale corpus, which favors SVM-based methods that have been trained to recognize a finite set of objects/concepts. The Video Google style task is more open-ended and, perhaps, more general -- one cannot enumerate the concepts or objects that a user might want to retrieve from video data. Instead, the user is asked to provide an example, and the most relevant matches are returned.

Nister and Stewenius evaluated their approach on a large scale (1M+ images) corpus, yet the bulk of the images were from copyright protected movies. The ground-truth subset, identified earlier as the NS data set, is approximately 10K of the 1M images, and can be freely downloaded. To reproduce their reported results on the large scale data set, one would have to extract the image frames for all seven movies identified in their paper, which is time-consuming and possibly in violation of digital rights management laws.

Oxford researchers have developed their own large scale benchmark consisting of 5K ground-truth images, gathered from Flickr, that show various Oxford landmark buildings. This Oxford 5K data set is embedded in approximately 1M Flickr images gathered using the top 450 most popular tags. The Oxford 5K is available for download, but not the entire large scale corpus. To reproduce their results, one has to build a new data set from the top Flickr tags, which will differ from the Oxford set because the source of the data is not static. So exact reproduction of results using either the NS or Oxford large scale data sets is impossible until the entire original data sets are made available.

Another issue with the Oxford data set is the focus on architectural landmarks, which represent rigid shapes with many coplanar surfaces. Evaluations on this data set, using a vocabulary generated from a similar set of images, may show a positive performance bias over a more general setup. \citet{philbin_lost_2008} report results on a vocabulary trained on Paris architectural landmarks used for \ir{} on the Oxford images. In general, it is unreasonable to expect that the visual vocabulary would be generated by images of the same thematic material as the query. Furthermore, verifying the query results using approximate affine homographies is likely to work well for query targets containing rigid objects with many coplanar features.

A somewhat newer data set is the INRIA Holidays images, also designed for the evaluation of large scale \ir{} systems \citep{jegou_hamming_2008}. Like the Oxford data, INRIA uses a set of ground truth images embedded in distractors culled from Flickr. To encourage comparative evaluation, the INRIA group provides data for the pre-computed features (Hessian Affine detector with SIFT descriptors) on 1 million Flickr images. The feature data is 235 gigabytes. This data allows comparative evaluation of million-image-scale \ir{} only for those approaches that use the same features.

Ideally, an evaluation benchmark would provide a variety of query images and would be able to mimic common use-cases for this technology. A technique that works best for searching an action movie for car wrecks and explosions may be less than ideal for organizing your vacation photos -- unless your vacations are substantially more exciting than ours! An evaluation benchmark should specify a neutral data set, testing protocol, and performance measures. Due to the costs in collecting, managing, and hosting terabyte-scale data sets, it may be best handled by an national or international standards body.

\section{Challenges}
\label{section:challenges}
Even though \bof{} representations have proven powerful for \ic{} and \ir{}, there remain challenges in applying them to other tasks. In this section, we describe a few of the limitations of the standard BoF representation, with the expectation that novel variants may yet emerge that mitigate these limitations while retaining the key strengths of the paradigm.

\subsection{Spatial Information}
The lack of spatial information in traditional BoF representations seems to make them poor choices for systems that localize objects in images or describe relationships among objects. Term vectors pool information from across an image, making it difficult to extract relational concepts such as ``a person standing next to a streetlight.'' Researchers have explored modifying BoF representations to encode some spatial information, for example Lazebnik et al.'s Spatial Pyramid Matching method discussed in Section \ref{section:IC}.  More spatial information might help with spatial tasks, such as finding small objects within cluttered scenes, or recognizing relations among objects. Stronger encodings of spatial information, however, move away from the BoF paradigm and toward part-based or constellation models, with the risk of loosing the simplicity and performance of BoF methods.

\subsection{Semantic Meaning}
While some visual words may represent distinct object parts, such as a headlight on a car, in practice most visual words have no simple linguistic description. This lack of inherent semantic meaning in a BoF representation poses a challenge for certain tasks, such as retrieving images from keywords or generating natural language descriptions from images. The nascent ImageNet database of Deng et al.  \citep{deng_imagenet:large-scale_2009} may be one approach to bridging the gap between visual and linguistic terms, although other, yet to be invented techniques may be needed also. 

\subsection{Performance Evaluation}
The observation that BoF representations pool information from across an image does more than make it difficult to localize targets, it makes it difficult to assess what BoF systems are actually recognizing. A good example comes from \citet{pinto_far_2009}. They built a system that achieves state-of-the-art face recognition performance on the complex \textit{Labeled Faces in the Wild} (LFW) data set. LFW contains images of celebrity faces culled from news photographs, and is considered a difficult face recognition data set because of the uncontrolled illumination and poses, and because of the complex backgrounds that surround the faces. Pinto et al. achieved their success using densely-sampled local image features with  Gabor-based feature  descriptors and a Multiple Kernel Learning (MKL) algorithm to match faces. Although not precisely a \bof{} model, it is similar to the BoF \ic{} approaches that use dense sampling and SVM classifiers. In a cautionary note, however, Pinto et al. suggest that their system's performance may have little to do with recognizing faces, and may instead be the result of exploiting low-level similarities among image backgrounds. They show that the same method fails to be robust to modest amounts of variation when tested on synthetic data. 

Pinto et al.'s observation is similar to one in Lazebnik et al. \citep{lazebnik_beyond_2006}. Lazebnik et al. note that certain categories in the Caltech101 benchmark, such as minarets, have corner artifacts induced from artificial image rotation. These artifacts are highly discriminative and stable cues that generate high performance on the CalTech data set, but do not generalize to unrotated pictures of minarets. Indeed, one would expect rotated images of other objects to be labeled as minarets! The challenge, therefore, is to evaluate BoF methods in such a way as to determine what, exactly, the system is recognizing, in order to know whether performance will generalize to future data sets. 

\section{Conclusion}
\label{section:conclusion}
The \bof{} representation is notable because of its relative simplicity and strong performance in a number of vision tasks. Nowak et al. demonstrated compelling state-of-the-art performance on the 2005 PASCAL Visual Object Recognition Challenge \citep{nowak_sampling_2006}. Nister and Stewenius developed break-through scalability by demonstrating \ir{} on a million-image data set \citep{nister_scalable_2006}. Continued research has addressed quantization issues, improved feature detectors and descriptors, and developed more compact representations and scalable indexing schemes. 

Yet challenges remain. Comparative evaluation of large-scale \ir{} tasks is difficult. This is ironic, because it is the success of the BoF representation that has led to the need for huge image data sets, the distribution of which is time and cost prohibitive. The sampling strategy remains an open issue. Is it better to use keypoint detectors or sample using a scale-space grid? Are SIFT-like descriptors the final answer to the features, or should color and other aspects be further explored? BoF methods provide little semantic description of image contents. This lack of semantic meaning makes it difficult to integrate BoF-based \ir{} with keyword text queries.

Finally, a BoF approach may be less suitable than other techniques for object detection and localization. One can imagine a ``Where's Waldo'' challenge, where a BoF method would misclassify an image as containing Waldo because the localized features have many horizontal red and white stripes. Because BoF representations do not identify and localize objects in images, object recognition performance can be questionable. An image is indicated as having an object because its ``bag'' contains the same type and distribution of features as other images containing the object. With no information on the arrangement of the features, many false detections could be expected in real-world applications.

As of this survey, BoF research is still very much an active field, and advances are published at major conferences every year. The technology has advanced considerably in the past decade, and we hope to see continued exploitation of this powerful and computationally cheap representation for a variety of applications.

\bibliographystyle{spbasic}
\bibliography{BoF}

\end{document}